\documentclass[11pt]{article}
\usepackage{smile}
\usepackage[colorlinks,
           linkcolor=blue,
           anchorcolor=blue,
           citecolor=blue
           ]{hyperref}

\usepackage{kpfonts}
\DeclareMathAlphabet{\mathsf}{OT1}{cmss}{m}{n}
\SetMathAlphabet{\mathsf}{bold}{OT1}{cmss}{bx}{n}

\usepackage{fullpage}

\begin{document}

\title{\huge Residual Network Based Direct Synthesis of EM Structures: A Study on One-to-One Transformers\thanks{keywords: direct synthesis; machine learning; rapid design}}

\author{
David Munzer{\textsuperscript{1}},
Siawpeng Er{\textsuperscript{2}},
Minshuo Chen{\textsuperscript{2}}, 
Yan Li{\textsuperscript{2}},\\\vspace{0.05in}
Naga S. Mannem{\textsuperscript{1}}, 
Tuo Zhao{\textsuperscript{2}}, 
Hua Wang{\textsuperscript{1}}
\vspace{0.07in}
\\
{\textsuperscript{1}}School of Electrical and Computer Engineering, Georgia Tech\\
{\textsuperscript{2}}School of Industrial and Systems Engineering, Georgia Tech\\
}

\date{}
\maketitle

\begin{abstract}
We propose using machine learning models for the direct synthesis of on-chip electromagnetic (EM) passive structures to enable rapid or even automated designs and optimizations of RF/mm-Wave circuits. As a proof of concept, we demonstrate the direct synthesis of a 1:1 transformer on a 45nm SOI process using our proposed neural network model. Using pre-existing transformer s-parameter files and their geometric design training samples, the model predicts target geometric designs.
\end{abstract}


\section{Introduction}

\begin{figure}[t]
	\centering
	\includegraphics[width=0.8\textwidth]{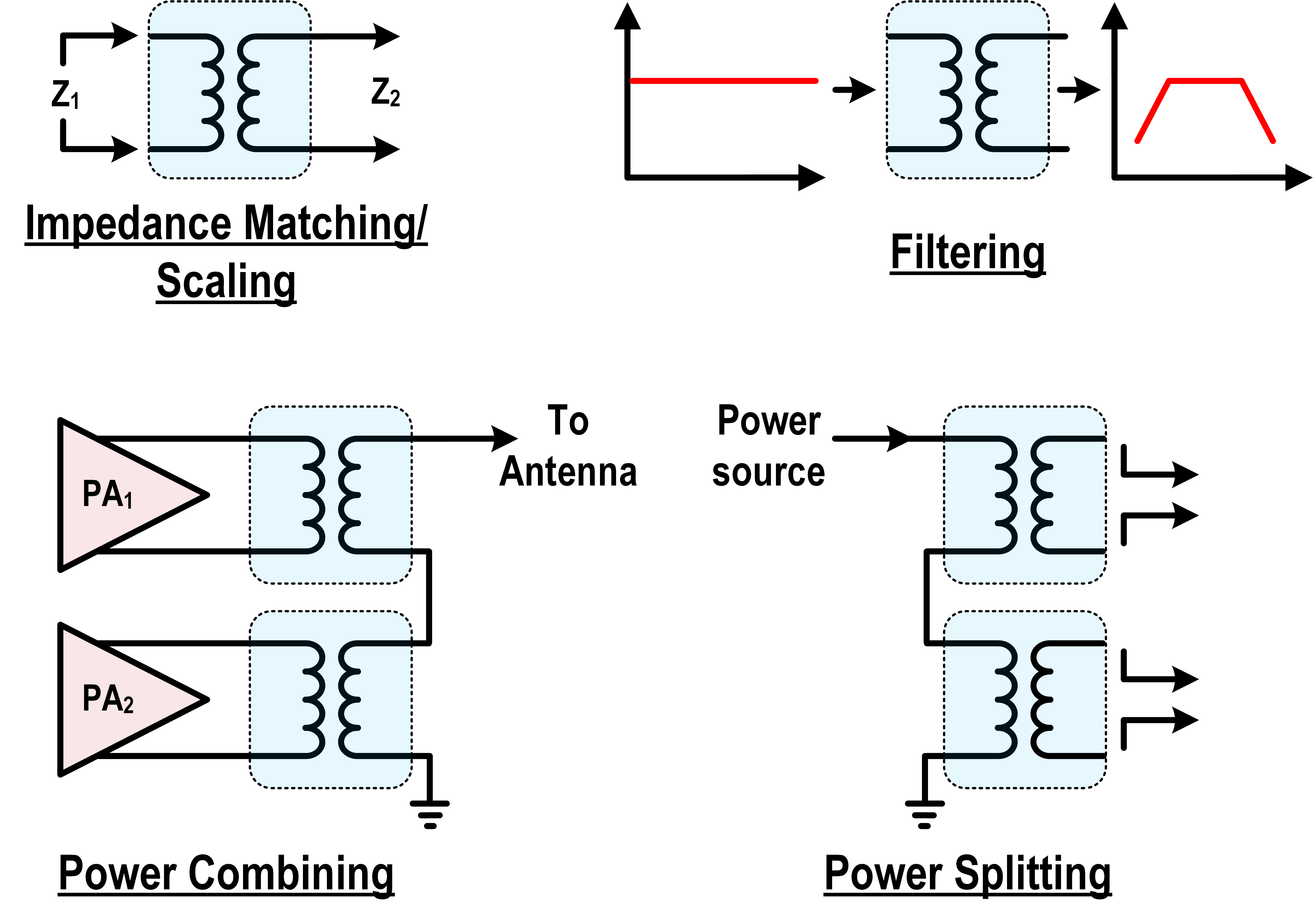}
	\caption{ General uses of transformers in RF/mm-Wave Design }
	\label{fig:design}
\end{figure}

RF/mm-Wave circuits are often governed by the design/ performance/ form factor of the passive components/networks used. Passives are extensively used for impedance matching, scaling, tuning, filtering, power combining/splitting, and signal generation (Figure \ref{fig:design}) Therefore, maximizing passive structures' performance while minimizing their form factor is critical for RF/microwave designs. However, the typical RF/mm-Wave design flow is very tedious: a designer will model the desired passive structure by using ideal elements such as lumped capacitors/inductors and transmission lines in the schematics. He / she needs to perform EM simulations to find passive structure geometries that can yield circuit performance matching that of the schematic model. This process is very iterative and requires extensive EM design background to arrive at a faster, optimal solution. The fundamental reason of this existing iterative and computationally inefficient design flow is that most existing EM simulation software suites only act as ``analysis tools'' (Figure \ref{fig:ml_design}). These software suites only analyze EM passive structures with given geometries and then yield their circuit performance parameters. To obtain the desired EM passive structures, designers need to analyze the corresponding EM simulation results and from theory/prior experience, change the geometrical parameters required to achieve their desired performance. The EM passive geometry with the best fit circuit performance is then selected and if a suitable geometry is not found, successive EM simulations or remodeling of the passives are required. This situation only worsens as the complexity of the passive structure increases. EM simulations using software such ANSYS's HFSS take a large amount of simulation time, which increases exponentially as the size and complexity of the passive structure increases. In addition, the number of geometrical parameters also increases which exponentially increases the possible number of solutions, further increasing the required number of iterations.

However, what designers actually need are ``synthesis tools'' which directly generate the passive geometries based on the required circuit specifications. Such EM synthesis tools will radically accelerate the design and optimization time of RF/mm-Wave circuits, reduce the dependency of expert-knowledge, and enable rapid, low-cost and knowledge-transferable RF/mm-Wave circuits.  It will also free designers from laborious iterations and implementations, allowing them to focus on topological or architectural innovations.

\section{EM Passive Structure Design Flow}

\begin{figure}[t]
	\centering
	\includegraphics[width=0.9\textwidth]{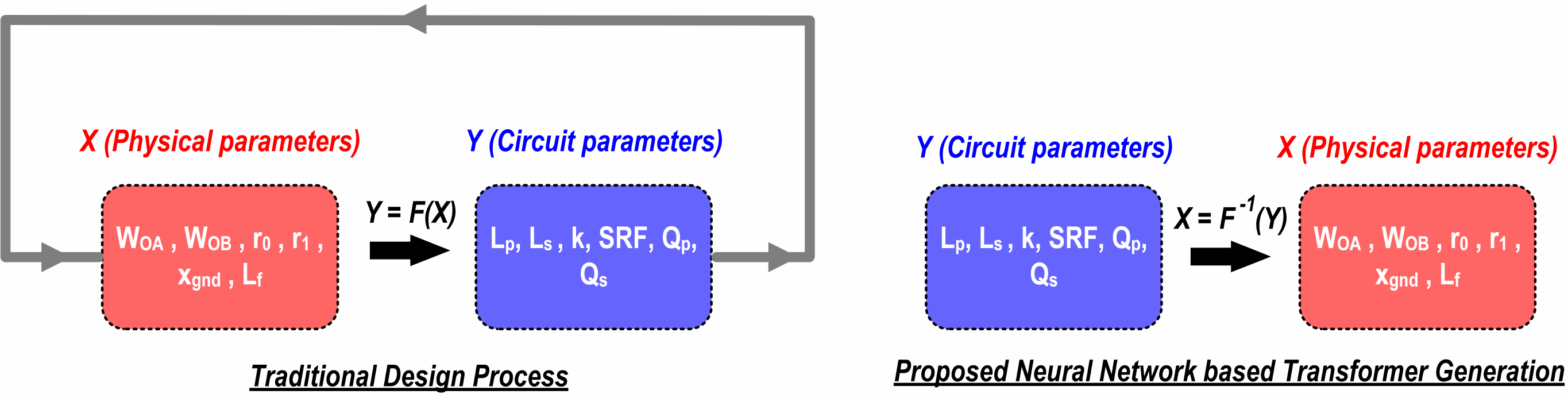}
	\caption{ Comparison of the existing iterative cycle for designers in which they iteratively tune their passive structures' geometry based on EM simulations and our neural network predictive model which gives the optimal geometry based on the desired circuit parameters.  }
	\label{fig:ml_design}
\end{figure}
\begin{figure}[t]
	\centering
	\includegraphics[width=0.8\textwidth]{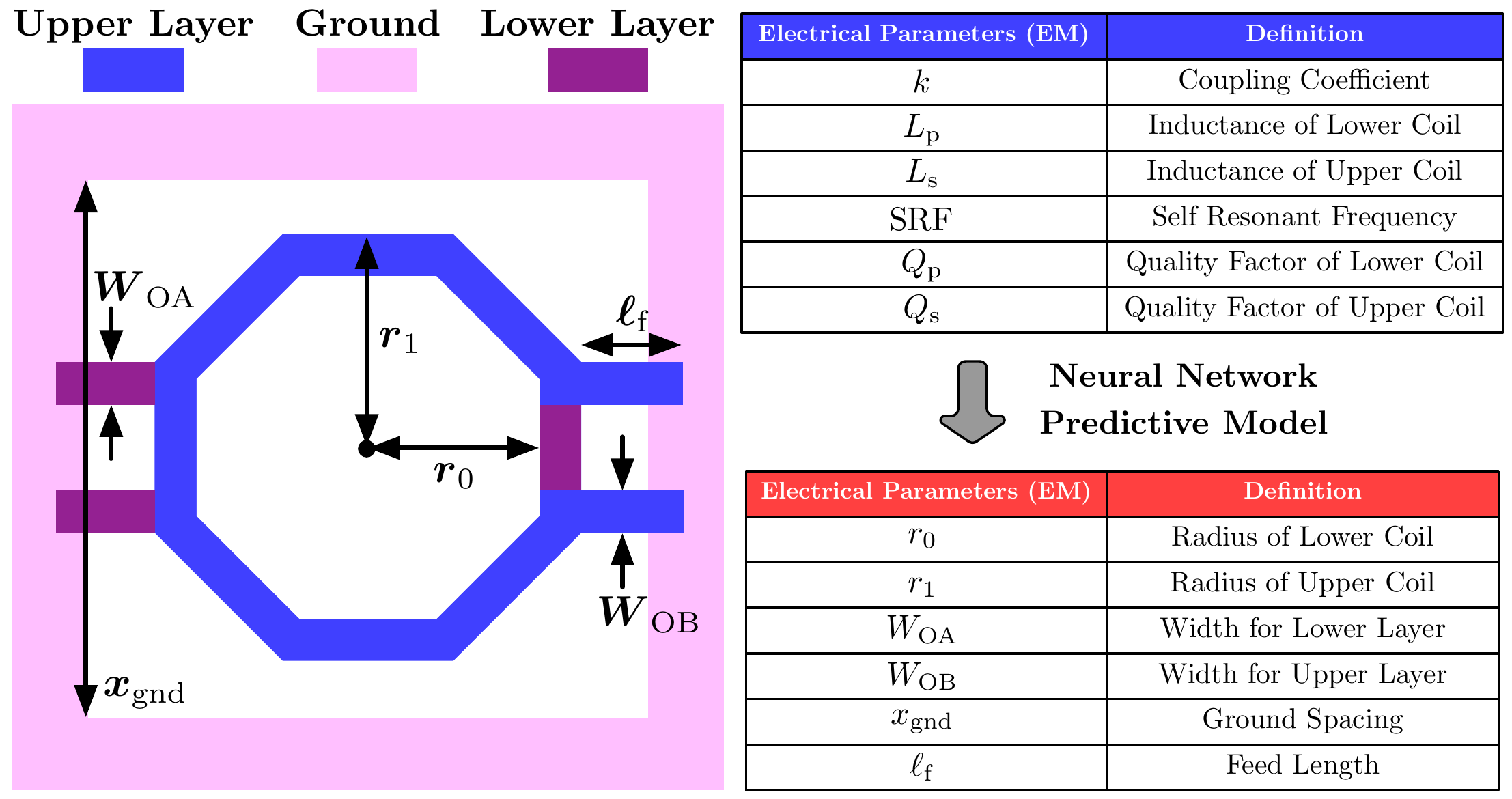}
	\caption{ EM transformer model with corresponding input circuit parameters and output geometrical parameters.  }
	\label{fig:parameters}
	
\end{figure}

On-chip transformers are used extensively in RF/mm-Wave designs, particularly for the upcoming 5G communication. They employ multiple roles ranging from impedance matching/scaling, differential to single ended conversion, filtering, biasing, power combining/dividing, etc. Specifically, 1:1 transformers are often used at mm-Wave due to their compact form factor, large achievable coupling, and broadband impedance transformation properties \cite{wang2019an}. Therefore, we propose a machine learning based predictive model (Figure \ref{fig:ml_design}) for the direct EM synthesis of 1:1 transformers, which will generate the desired transformer design parameters, including the coil radiuses ($r_0$ and $r_1$), widths  ($W_\textrm{OA}$ and $W_\textrm{OB}$), ground spacing ($x_\textrm{gnd}$), and input/output feed length ($\ell_\textrm{f}$), based on the targeted circuit parameters including the self-resonance frequency (SRF), primary and secondary inductance ($L_\textrm{p}$ and $L_\textrm{s}$), coupling coefficient ($k$), and primary/secondary quality factor ($Q_\textrm{p}$ and $Q_\textrm{s}$). See Figure \ref{fig:parameters} for the synthesized transformer structure, input parameters, and output geometrical parameters.

Our EM predictive model is built upon residual network architectures. Neural networks are known for their predictive power: they can provide a highly close fit to new data after training. Empirical results also suggest that overparameterized neural networks (the number of free parameters exceeds that of training data points) are easy to train, and surprisingly retain appealing predictive performance. More recently, residual networks \cite{he2016deep} further ease the training, and enhance the prediction by allowing direct interactions between inputs and outputs. We train the model using a limited number of transformers s-parameter files and their geometric designs. 
\textit{We use s-parameter files instead of measuring fabricated transformers since doing so would only verify the accuracy of the EM simulator instead of our algorithm}.
When given targeted electrical parameters, the neural network outputs geometric designs which act as a starting point to close in on an optimal solution, hence acting as the ``EM synthesis tool''. Thus, the designers only need very few additional EM simulations to verify and fine tune the design parameters, which circumvents the tedious and resource intensive, iterative process.

\section{Residual Network Architecture and Training}
\begin{figure}[t]
	\centering
	\includegraphics[width=0.8\textwidth]{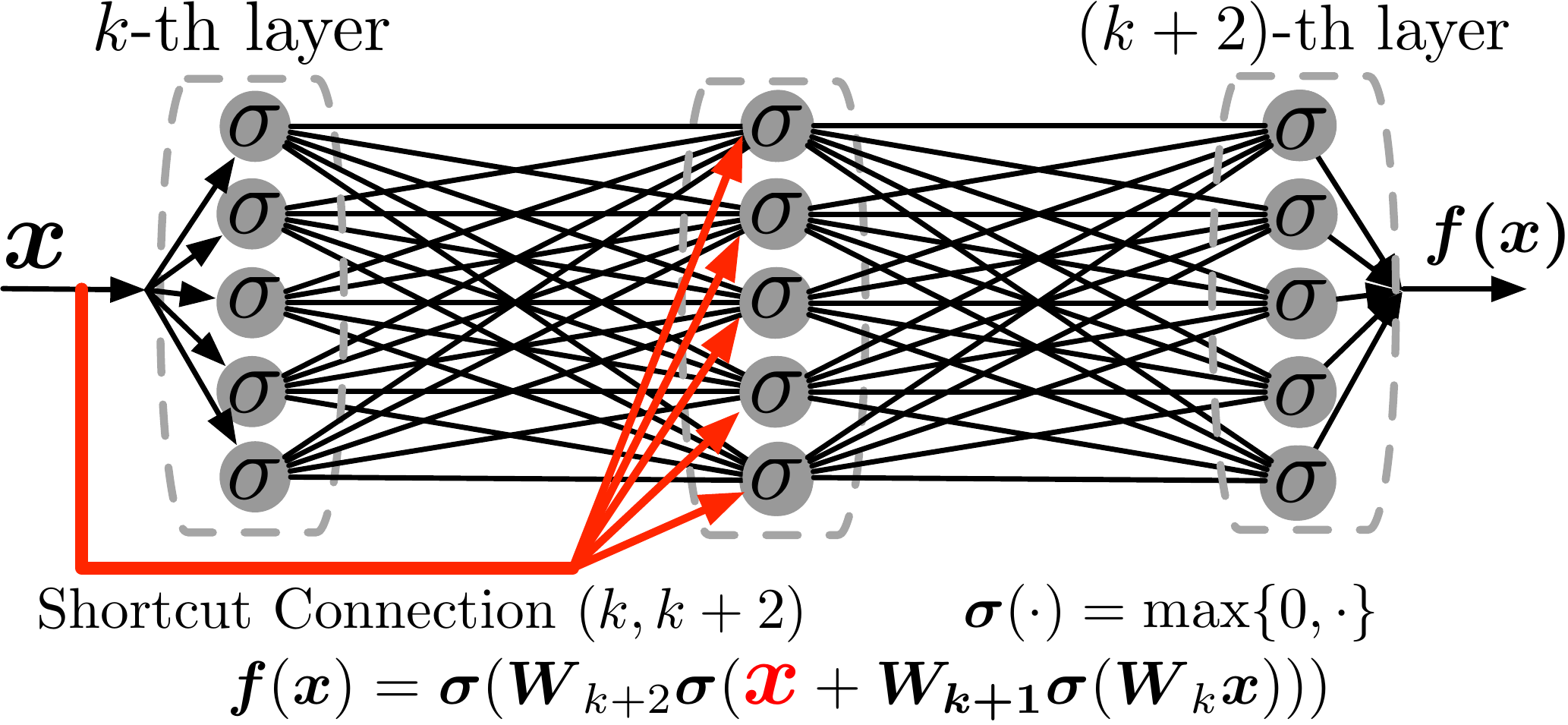}
	\caption{Illustration of residual block with shortcut connection $(k,k+2)$. The input of the $k$-th layer is directly added to the input of the $(k+2)$-th layer.   }
	\label{fig:resnet}
\end{figure}

The residual network architecture consists of a series of residual blocks. Each residual block is built upon a feedforward neural network by adding shortcut connections across layers. Figure \ref{fig:resnet} illustrates a residual block with a shortcut connection bypassing two hidden layers. Such a residual block essentially consists of two sub-networks (feedforward neural networks) of different complexities: one is more complex (black, two-layer) and the other is simpler (red). By concatenating multiple residual blocks, we generate a collection of sub-networks with different complexities. Thus, residual neural networks can be viewed as an ensemble of feedforward neural networks containing varying and adaptive numbers of layers. This largely improves the modeling ability of residual networks. More importantly, for general feedforwad neural networks, simply stacking layers does not promise a performance boost. One of the major reasons is that the vanishing/exploding gradient issue arises which makes the network  very difficult to train \cite{Glorot10understandingthe}. However, the residual network architecture mitigates this error through the shortcut connections across layers with additional performance boosts.

The training of neural networks can be written as minimizing the following penalized empirical loss:
\begin{align}\label{eq:nntrain}
\textstyle
\min_{\theta}~ L(\theta) = \Phi_n\left(\{f_\theta(x_i), y_i\}_{i=1}^n\right) + R(\theta),
\end{align}
where $f_\theta(x_i)$ is the predicted output of neural network models, i.e., $1$:$1$ transformer's geometric parameters, in this case.
Here $\theta$ denotes the weight parameters, $\Phi_n$ is a properly chosen loss function and the subscript $n$ emphasizes the dependence on $n$ samples, and $R$ is a penalty to avoid overfitting. Here $(x_i, y_i)$'s are samples with $x_i$ denoting input (circuit parameters) and $y_i$ denoting targeted response (geometric parameters), and $n$ is the sample size. In practice, $\Phi_n$ are often chosen as an average of empirical errors, e.g., $\Phi_n = \frac{1}{n} \sum_{i=1}^n (f_\theta(x_i) - y_i)^2$ corresponds to the mean squared error.

We can apply the gradient descent algorithm to minimize \eqref{eq:nntrain}, i.e., at each iteration, $\theta \leftarrow \theta - \eta \nabla_\theta L(\theta)$ with $\eta$ being the learning rate. Due to the special structure of neural networks (consisting of multiple layers), we can use backpropagation to calculate the gradient $\nabla_\theta L$ \cite{lecun1990handwritten} . 

Exact evaluation of $\nabla_\theta L$, however, is computationally intensive especially when $n$ is large and the neural network is highly complex. A common practice is to apply stochastic gradient descent (SGD) type algorithms. At each iteration, we randomly select a small number of training samples (i.e., mini-batch) to calculate a stochastic approximation of $\nabla_\theta L$:
\begin{align*}
\textstyle
\nabla_\theta L(\theta) \approx \widehat{\nabla}_\theta L(\theta) = \nabla_\theta \Phi_b\left(\{f_\theta(x_{i}), y_{i}\}_{i \in \mathcal{B}}\right) + \nabla_\theta R(\theta),
\end{align*}
where $b \ll n$ is the size of mini-batch $\mathcal{B}$.


\section{1:1 On-chip Transformer Direct Synthesis Demonstration}

\subsection{Experiment Setup}
We evaluate our predictive models on a $1$:$1$ transformer design task using three residual network architectures. We also compare the predictive models with three baseline methods: linear regression (LR), gradient boosting (GB) \cite{hastie2005elements}, and feedforward neural network (FN). The network configurations are listed in Table \ref{tab:architecture}. Note that feedforward neural network models FN$_i$ and residual network models $\mathcal{N}_i$ for $i = 5, 6, 7$ have the same total number of parameters.

\begin{table}[t]
	\caption{Feedforward and Residual Neural Network Architecture}
	\label{tab:architecture}
	\begin{center}
		\begin{tabular}{| @{ }c@{ } || @{ }c@{ } | @{ }c@{ } |}
			\hline
			Model & Width of Hidden Layers & Shortcut Connections \\
			\hline
			FN$_i$ & $\{2048 \times i\},  i=2, \ldots ,7$ & NA \\
			\hline
			&& \\[-1em]
			$\mathcal{N}_5$ & $\{2048 \times 5\}$ & (1,5) \\
			\hline
			&& \\[-1em]
			$\mathcal{N}_6$ & $\{2048 \times 6\}$ & (1,6) \\
			\hline 
			&& \\[-1em]
			$\mathcal{N}_7$ & $\{2048 \times 7\}$ & (1,3), (3,5), (5,7) \\
			\hline
		\end{tabular}
	\end{center}
\end{table}

During the training, geometric and circuit parameters of randomly selected pre-solved on-chip $1$:$1$ transformer designs are used as the input data, which we standardize before feeding to the neural networks.
We use Adam \cite{kingma2014adam} as our optimizer, one of the most widely used SGD type algorithms for training neural networks. Adam enjoys faster convergence in practice by using adaptive learning rate and momentum acceleration. 
The details of the algorithm are presented in Algorithm \ref{alg:adam}.
In our experiments, we randomly select $16$ samples from the training set at each iteration to form our mini-batch.
\begin{algorithm}[h]
	\caption{ADAM algorithm, $\sqrt{\cdot}$, $(\cdot)^{-1}$, and $\odot$ denote element-wise square root,  inverse,  and  multiplication.}
	\label{alg:adam}
	\begin{algorithmic}
		\STATE{\textbf{Input:} 
			learning rate $\eta$, $\beta_1, \beta_2 $},  $\epsilon $, weight decay $w$.
		\STATE{\textbf{Initialize:} $\theta_0$, $m_0 = 0$, $v_0 = 0$, $t = 0$.}
		\WHILE{$\theta_t$ not converged}
		\STATE{ Set $t = t+1$, choose mini-batch  $\mathcal{B} \subset \{1, \ldots, n\}$}
		\STATE{$g_t = \nabla_\theta \Phi_b \left(\{f_{\theta_{t-1}}(x_i), y_i\}_{i \in \mathcal{B}}\right)$ with $b = |\mathcal{B}|$}
		\STATE{$m_t = \beta_1 m_{t-1} + (1-\beta_1) g_t$ }
		\STATE{$v_t = \beta_2 v_{t-1} + (1-\beta_2) g_t\odot g_t$}
		\STATE{$\hat{m}_t =  m_t / (1 - \beta_1^t)$, $\hat{v}_t = v_t / (1 - \beta_2^t) $}
		\STATE{$\theta_t = (1 -\eta w) \theta_{t-1} - \eta \hat{m}_t \odot (\sqrt{\hat{v}_t} + \epsilon )^{-1} $ }
		\ENDWHILE
		\RETURN{ $\theta_t$}
	\end{algorithmic}
\end{algorithm}

\subsection{Model Comparisons}\label{sec:ex_N_3}
We perform extensive comparisons on the prediction accuracy of different models.
We use EM simulators to obtain $6400$ pairs of $1$:$1$ physical transformer parameters and their corresponding circuit parameters. 
We randomly select a testing set consisting of $1200$ samples, and vary the size of training set in $\{600, 1200, 2400, 4800\}$.

%
%
%
%
%
%
We use two different training loss metrics: 1) Scaled Mean Squared Error (SMSE)
\begin{align}
\textstyle
\textrm{SMSE}(\theta) = \frac{1}{nk} \sum_{i = 1}^n \sum_{j = 1}^k \left(\frac{y_{i,j}  - \hat{y}_{i,j}(\theta)}{ y_{i,j}} \right)^2.
\end{align}
2) Scaled Dimensional Mean Squared Error (SDMSE) \cite{liu2015calibrated}
\begin{align}
\textstyle
\textrm{SDMSE}(\theta) = \frac{1}{k} \sum_{j = 1}^k  \sqrt{\frac{1}{n}\sum_{i = 1}^n\left(\frac{y_{i,j}  - \hat{y}_{i,j}(\theta)}{ y_{i,j}} \right)^2}.
\end{align}
In the above, $\hat{y}(\theta) = f_\theta(x_i)$ denotes the predicted geometrical parameters (totally $k$ parameters). Note that SMSE minimizes the relative error of each prediction, which accounts for the different scales of physical parameters and stabilizes the training. Moreover, SDMSE puts additional emphasis on balancing the prediction error across testing samples.

Our training objective $L(\theta)$ is obtained by incorporating the weight decay penalty, i.e., $R_w(\theta) = \frac{w}{2}\|\theta\|_2^2$:
\begin{align}\label{eq:obj}
L(\theta) = \Phi_n\left(\{f_\theta(x_i), y_i\}_{i=1}^n \right) + R_w(\theta),
\end{align}
where $\Phi_n$ takes SMSE or SDMSE defined before, and $w$ controls the strength of weight decay.

For all experiments, we set the  $w =10^{-4}$ in \eqref{eq:obj}. In fact, the prediction accuracy is not sensitive to $w$, since we only observe negligible difference by fine tuning $w$.
We repeat $30$ independent experiments for each predictive model and report the average SMSE and $R^2$-score on the testing set. $R^2$-score is commonly used for indicating goodness of fit. Different from the relative error-based metric SMSE, $R^2$-score is calculated based on the proportion of total variation of geometrical parameters in the testing set explained by the predictive model.




\begin{table}[t]
	\caption{Performance of Predicting Geometrical Parameters using Circuit Parameters}
	\label{tab:experiments}
	\begin{center}
		\begin{tabular}{ | c || c | c || c | c |}
			\hline
			\multirow{2}{*}{Model}  & \multicolumn{2}{c ||}{SMSE Training Loss} & \multicolumn{2}{c|}{SDMSE Training Loss} \\ 
			&&&& \\[-1em]
			& SMSE & $R^2$ & SMSE & $R^2$ \\
			\hline
			&&&& \\[-1em]
			LR & 0.0358 & 0.5620 & 0.0358 & 0.5620 \\
			\hline
			&&&& \\[-1em]
			GB & 0.0199 & 0.7468 & 0.0199 & 0.7468 \\
			\hline
			&&&& \\[-1em]
			FN$_2$ & 0.0114  &  0.7553 & 0.0054 & 0.9070 \\
			\hline
			&&&& \\[-1em]
			FN$_3$ & 0.0070  &  0.8457  & 0.0040 & 0.9360 \\
			\hline
			&&&& \\[-1em]
			FN$_4$ & 0.0065  &  0.8410  & 0.0038 & 0.9400 \\
			\hline
			&&&& \\[-1em]
			FN$_5$ & 0.0061 & 0.8183  & 0.0037 & 0.9431 \\
			\hline
			FN$_6$ & 0.0079 & 0.8040  & 0.0036 & 0.9432 \\
			\hline
			&&&& \\[-1em]
			FN$_7$ & 0.0080 & 0.8061  & 0.0037 & 0.9051 \\
			\hline
			&&&& \\[-1em]
			$\mathcal{N}_5$ & 0.0051  & 0.8846  &  0.0030  & 0.9535  \\
			\hline
			&&&& \\[-1em]
			$\mathcal{N}_6$ & 0.0049   &  0.8868 &  0.0031 &  0.9553 \\
			\hline
			&&&& \\[-1em]
			$\mathcal{N}_7$ & 0.0043 & 0.9243  & 0.0030 & 0.9586 \\
			\hline
		\end{tabular}
	\end{center}
\end{table}

We observe that residual networks consistently outperform other models on both evaluation criteria, when varying the size of training set. We summarize the experimental results corresponding to using $2400$ training samples in Table \ref{tab:experiments}.

For all neural network models, using SDMSE as training loss improves the testing accuracy compared to SMSE.
It is shown that residual neural networks yield superior performance compared to feedforward neural networks consisting of the same number of weight parameters: consistently better prediction accuracy and less sensitivity to the training loss we choose.
In addition, adding more layers to the feedforward network does not improve the prediction accuracy.
The feedforward network  FN$_7$ shows substantial performance degradation compared  to its shallow counterparts FN$_5$ and FN$_6$.
These observations indicate that shortcut connections play a crucial role in the superior performance  of residual networks.

We also observe that
$\mathcal{N}_7$ consistently achieves the highest prediction accuracy. It is worth mentioning that the performance gap between residual networks $\mathcal{N}_7$ and $\mathcal{N}_5$ or $\mathcal{N}_6$ are more significant when we use SMSE as the training loss.
Note that $\mathcal{N}_5$ and  $\mathcal{N}_6$ both contain only one shortcut connection linking the input layer directly to the output layer, while $\mathcal{N}_7$ is equipped with more sophisticated shortcut connections.
This observation indicates that a careful design of shortcut connections in $\mathcal{N}_8$ can achieve a significant performance boost.

%

\begin{table}[t]
	\caption{Performance of $\mathcal{N}_7$ using Different Training Sizes}
	\label{tab:trainingsize_experiments}
	\begin{center}
		\begin{tabular}{ | c || c | c || c | c |}
			\hline
			\multirow{2}{*}{Training Size}  & \multicolumn{2}{c ||}{With Feed Length} & \multicolumn{2}{c|}{Without Feed Length} \\ 
			&&&& \\[-1em]
			& SMSE & $R^2$ & SMSE & $R^2$ \\
			\hline
			&&&& \\[-1em]
			600 & 0.0090  &  0.8940  & 0.0028 & 0.9217 \\
			\hline
			&&&& \\[-1em]
			1200 &  0.0052 & 0.9337 & 0.0024 & 0.9433 \\
			\hline
			&&&& \\[-1em]
			2400 & 0.0033 & 0.9586 & 0.0016 & 0.9593 \\
			\hline
			&&&& \\[-1em]
			4800 & 0.0022 & 0.9666 & 0.0018 &  0.9670 \\
			\hline
		\end{tabular}
	\end{center}
\end{table}
\begin{table}[t]
	\caption{SMSE for Predicting Each Geometrical Parameter}
	\label{tab:smse_each}
	\begin{center}
		\begin{tabular}{| @{ }c@{ } || @{ }c@{ } | @{ }c@{ } | @{ }c@{ } | @{ }c@{ } | @{ }c@{ } | @{ }c@{ } |}
			\hline
			Geometrical Parameter & $W_\textrm{OA}$  & $W_\textrm{OB}$ & $r_0$ & $r_1$ & $x_\textrm{gnd}$ & $\ell_\textrm{f}$ \\
			\hline
			&&&&&& \\[-1em]
			SMSE & 0.0017 & 0.0038 & 0.0003 & 0.0007 & 0.0012 &  0.0123 \\
			\hline
		\end{tabular}
	\end{center}
\end{table}
\vspace{- 0.15 in}

\subsection{Further Experiments on Residual Networks}
We further present more comprehensive experimental results for the residual network $\mathcal{N}_7$.
As we have observed in Table \ref{tab:experiments}, using SDMSE as the training loss improves the prediction accuracy compared to SMSE. Thus, we focus on SDMSE loss with weight decay.
The setup of the experiment is exactly the same as that  in Section \ref{sec:ex_N_3}.

The results of using different sizes of training set are summarized in Table \ref{tab:trainingsize_experiments}. 
We see that as the size of the training set increases, the prediction accuracy of $\mathcal{N}_7$ also improves.

Moreover, we evaluate the prediction power of the residual network $\mathcal{N}_7$ on each geometrical parameter. 
Table \ref{tab:smse_each} reports the SMSE for predicting each geometrical parameter , when using a training size of $2400$ and SDMSE as the training loss.

It can be seen that the SMSE for predicting the feed length $(\ell_{\textrm{f}})$ well exceeds those for predicting other geometrical parameters. This observation is consistent across different training sizes and both SMSE and SDMSE training loss.
The low correlation between the circuit parameters and the feed length well matches theory, since feed length only influences the inductance of the primary/secondary coils. In addition, its choice is largely independent of the transformer geometric design, whereas highly relies on the physical layout of the RF/mm-Wave circuit.

Therefore, we further test using $\mathcal{N}_7$ to predict all the geometrical parameters except the feed length $(\ell_{\textrm{f}})$.
The setup of the experiment is exactly the same as in section \ref{sec:ex_N_3}. The results are summarized in the rightmost two columns of Table \ref{tab:trainingsize_experiments}. By removing the feed length parameter, $\mathcal{N}_7$ enjoys a performance boost especially using a small number of training samples. This result is inspiring and suggests that $\mathcal{N}_7$ is indeed efficient in capturing the informative correspondence between circuit and geometrical parameters.

%

\subsection{Validation Examples of the Direct Synthesis}
We demonstrate an example of using the trained predictive model $\mathcal{N}_7$ to directly synthesize geometrical parameters given a randomly selected set of desired circuit parameters.
The obtained geometrical parameters are shown in Table \ref{tab:predict_experiments}. 
Then, we run one EM simulation using the predicted geometry to verify the prediction.
We observe that the synthesized circuit parameters closely match the desired parameters.
\begin{table}[H]
	\caption{Predicting Geometrical Parameters using Circuit Parameters}
	\label{tab:predict_experiments}
	\begin{center}
		\begin{tabular}{|c | c || c | @{ }c@{ } | @{ }c@{ } | @{ }c@{ } | @{ }c@{ } | @{ }c@{ } |}
			\hline
			\multicolumn{2}{|c ||}{Circuit Prameters} & $L_\textrm{p}$(pH) & $L_\textrm{s}$(pH) & $k$ & $\textrm{SRF}$(GHz) & $Q_\textrm{p}$ & $Q_\textrm{s}$ \\ \hline
			\multirow{2}{*}{I} & Targeted & 142.25 & 163.60 & 0.55 & 97.00 & 22.20 & 20.52 \\ 
			& Synthesized & 142.91 & 164.41 & 0.56 & 96.50 & 22.39 & 20.42 \\\hline
			\multirow{2}{*}{II} & Targeted & 173.30 & 188.44 & 0.48 & 99.00 & 21.81 & 23.59 \\ 
			& Synthesized & 168.48 & 184.43 & 0.47 & 99.00 & 21.89 & 24.36 \\\hline
			\multirow{2}{*}{III} & Targeted & 226.89 & 242.25 & 0.70 & 66.30 & 22.38 & 21.44 \\ 
			& Synthesized & 236.02 & 252.69 & 0.69 & 65.00 & 22.79 & 21.80 \\\hline
			\multirow{2}{*}{IV} & Targeted & 111.26 & 128.72 & 0.59 & 95.80 & 23.25 & 19.97 \\ 
			& Synthesized & 111.93 & 129.38 & 0.59 & 96.20 & 24.03 & 20.00 \\\hline
			\multirow{2}{*}{V} & Targeted & 245.09 & 294.89 & 0.62 & 78.70 & 21.79 & 16.73 \\ 
			& Synthesized & 243.21 & 293.00 & 0.62 & 79.00 & 21.78 & 16.84 \\\hline
		\end{tabular}
		
	\end{center}
	
	\vspace{0.07in}
	
	\begin{center}
		\begin{tabular}{|@{ }c@{ } || @{ }c@{ } | @{ }c@{ } |}
			\hline
			\multirow{3}{*}{Synthesized Geometry I}& $W_{\textrm{OA}} = 10.05 \mu$m & $r_0 = 45.32\mu$m \\
			& $W_{\textrm{OB}} = 9.98\mu$m & $r_1 = 52.24\mu$m \\
			& $x_{\textrm{gnd}} = 60.74\mu$m & $\ell_{\textrm{f}} = 24.03\mu$m \\\hline
			\multirow{3}{*}{Synthesized Geometry II}& $W_{\textrm{OA}} = 4.98 \mu$m & $r_0 = 41.32\mu$m \\
			& $W_{\textrm{OB}} = 7.99\mu$m & $r_1 = 50.93\mu$m \\
			& $x_{\textrm{gnd}} = 67.93\mu$m & $\ell_{\textrm{f}} = 28.73\mu$m \\\hline
			\multirow{3}{*}{Synthesized Geometry III}& $W_{\textrm{OA}} = 10.01 \mu$m & $r_0 = 62.38\mu$m \\
			& $W_{\textrm{OB}} = 9.99\mu$m & $r_1 = 67.46\mu$m \\
			& $x_{\textrm{gnd}} = 78.04\mu$m & $\ell_{\textrm{f}} = 14.98\mu$m \\\hline
			\multirow{3}{*}{Synthesized Geometry IV}& $W_{\textrm{OA}} = 14.93 \mu$m & $r_0 = 44.79\mu$m \\
			& $W_{\textrm{OB}} = 11.83\mu$m & $r_1 = 46.97\mu$m \\
			& $x_{\textrm{gnd}} = 60.64\mu$m & $\ell_{\textrm{f}} = 15.00\mu$m \\\hline
			\multirow{3}{*}{Synthesized Geometry V}& $W_{\textrm{OA}} = 4.97 \mu$m & $r_0 = 54.75\mu$m \\
			& $W_{\textrm{OB}} = 2.00\mu$m & $r_1 = 54.95\mu$m \\
			& $x_{\textrm{gnd}} = 68.90\mu$m & $\ell_{\textrm{f}} = 25.21\mu$m \\\hline
		\end{tabular}
	\end{center}
\end{table}


\section{Conclusion}
We propose a neural network based model for the direct synthesis of RF/mm-Wave EM passive structures. A proof of concept is  demonstrated on a $1$:$1$ transformer. Our trained residual network model generates near perfect predictions on transformer's geometrical parameters for given target circuit performance, and outperforms widely used machine learning baseline methods. Our proposed model can be further extended to more complex EM passive structures and revolutionize  the design procedure and automation of RF/mm-Wave circuits.

\bibliographystyle{ieeetr}
\bibliography{references}

\end{document}